%% file: _main.tex
\documentclass[conference]{IEEEtran}
\IEEEoverridecommandlockouts

\usepackage{cite}
\usepackage{amsmath,amssymb,amsfonts}
\usepackage{algorithmic}
\usepackage{graphicx}
\usepackage{textcomp}
\usepackage{xcolor}
\usepackage{url}

\usepackage{tcolorbox}
\usepackage{caption}
\usepackage{enumerate}
\usepackage{booktabs}
\usepackage{multirow}
\usepackage{enumitem}
\usepackage{caption}
\usepackage{subcaption}
\usepackage{makecell} 
\usepackage{tikz}

\newcommand{\quotes}[1]{``#1''}

\definecolor{fundo}{HTML}{CDCDCD}
\definecolor{preto}{HTML}{1C2127}
\definecolor{datadescription}{HTML}{FF4B3E}
\definecolor{featuredescription}{HTML}{FF3E61}
\definecolor{summarystatisticscorr}{HTML}{FFBD3D}
\definecolor{summarystatisticspps}{HTML}{F1FF3D}
\definecolor{summarystatisticsna}{HTML}{42DB63}
\definecolor{expectedoutput}{HTML}{3DCBFF}
\definecolor{baselines}{HTML}{443DFF}
\definecolor{shap_abs}{HTML}{833DFF}
\definecolor{split}{HTML}{631E18}
\definecolor{permutation}{HTML}{F2AC00}
\definecolor{shapquartiles}{HTML}{006E1A}
\definecolor{shapspearman}{HTML}{3D63D0}
\definecolor{auxmetrics}{HTML}{381673}

\newcommand\copyrighttext{%
  \footnotesize \textcopyright 2025 IEEE. Personal use of this material is permitted.
  Permission from IEEE must be obtained for all other uses, in any current or future
  media, including reprinting/republishing this material for advertising or promotional
  purposes, creating new collective works, for resale or redistribution to servers or
  lists, or reuse of any copyrighted component of this work in other works.}
\newcommand\copyrightnotice{%
\begin{tikzpicture}[remember picture,overlay]
\node[anchor=south,yshift=10pt] at (current page.south) {\fbox{\parbox{\dimexpr\textwidth-\fboxsep-\fboxrule\relax}{\copyrighttext}}};
\end{tikzpicture}%
}

\def\BibTeX{{\rm B\kern-.05em{\sc i\kern-.025em b}\kern-.08em
    T\kern-.1667em\lower.7ex\hbox{E}\kern-.125emX}}
\begin{document}

\title{Leveraging Large Language Models for Tacit Knowledge Discovery in Organizational Contexts\\
\thanks{This work was funded by the authors' individual grants from Kunumi.}
}

\author{
\IEEEauthorblockN{
    Gianlucca Zuin\IEEEauthorrefmark{1}\IEEEauthorrefmark{3}, 
    Saulo Mastelini\IEEEauthorrefmark{2}\IEEEauthorrefmark{3}, 
    Túlio Loures\IEEEauthorrefmark{1}\IEEEauthorrefmark{3}, 
    Adriano Veloso\IEEEauthorrefmark{1}\IEEEauthorrefmark{3}
}
\IEEEauthorblockA{\IEEEauthorrefmark{1}Universidade Federal de Minas Gerais (UFMG), Brazil}
\IEEEauthorblockA{\IEEEauthorrefmark{2}Instituto de Ciências Matemáticas e de Computação, Universidade de São Paulo (ICMC-USP), Brazil}
\IEEEauthorblockA{\IEEEauthorrefmark{3}Kunumi, Brazil}
\IEEEauthorblockA{Email: \{gianlucca, saulo, tulio, adriano\}@kunumi.com}
}



\maketitle
\copyrightnotice

\begin{abstract}

Documenting tacit knowledge in organizations can be a challenging task due to incomplete initial information, difficulty in identifying knowledgeable individuals, the interplay of formal hierarchies and informal networks, and the need to ask the right questions. To address this, we propose an agent-based framework leveraging large language models (LLMs) to iteratively reconstruct dataset descriptions through interactions with employees. Modeling knowledge dissemination as a Susceptible-Infectious (SI) process with waning infectivity, we conduct 864 simulations across various synthetic company structures and different dissemination parameters. Our results show that the agent achieves 94.9\% full-knowledge recall, with self-critical feedback scores strongly correlating with external literature critic scores. We analyze how each simulation parameter affects the knowledge retrieval process for the agent. In particular, we find that our approach is able to recover information without needing to access directly the only domain specialist. These findings highlight the agent’s ability to navigate organizational complexity and capture fragmented knowledge that would otherwise remain inaccessible.

\end{abstract}




\input{Text/1intro}

\input{Text/2related}

\input{Text/3method}
\input{Text/4exp}
\input{Text/5discussion}
\input{Text/6conclusion}


\section*{Code and Data Availability}
\label{sec:availability}
The code, data, and prompts used for the machine-learning analyses, available for non-commercial use, has been deposited at \url{https://doi.org/10.6084/m9.figshare.28785524}~\cite{figsharecode2025}.


\bibliographystyle{IEEEtran}
\bibliography{bibfile}

\end{document}

%% file: Text/1intro.tex
\section{Introduction}

In current fast-paced and knowledge-driven organizations, efficient management and sharing of knowledge remain challenging tasks. While explicit knowledge is easily documented, tacit knowledge — rooted in personal experiences and expertise — often proves difficult to systematically capture and disseminate. Sociological theories of organizational structure, including Max Weber's bureaucratic model~\cite{webertheory} and Frederick Taylor's scientific management~\cite{taylor2004scientific}, provide valuable insights into how both formal and informal systems influence the flow of information. Weber emphasized the efficiency of rigid hierarchies in structuring authority and information flow, while Taylor focused on the benefits of standardizing knowledge to boost productivity. However, both perspectives highlight inherent challenges: rigid hierarchies can create bottlenecks, while more flexible systems risk fostering knowledge silos.


Organizational structures, whether hierarchical, flat, or matrixed, impact how knowledge is disseminated within a company. These structures define the pathways of interaction among employees and influence knowledge's accessibility and retention. Research has shown that knowledge dissemination in cooperative learning networks resembles the spread of disease through a crowd~\cite{kiss2010can, zhu2015tacit}. 
Thus, by drawing on sociology and network theory, we illustrate information flow within organizations using epidemic models. The works of Polanyi~\cite{polanyi2009tacit} and Nonaka and Takeuchi~\cite{nonaka2007knowledge} further underline the complexity of capturing and sharing tacit knowledge. Polanyi's assertion that \quotes{we can know more than we can tell} underscores the implicit nature of such knowledge, while Nonaka and Takeuchi's SECI model demonstrates how organizations could convert tacit knowledge into explicit forms through socialization, externalization, combination, and internalization.

On the other hand, Large Language Models (LLMs) have demonstrated remarkable capabilities across a variety of tasks, including natural language understanding, translation, summarization, and creative writing~\cite{hadi2024large}. Their proficiency in processing and generating human-like text has opened new possibilities in areas such as automated content creation, human-computer interaction, and educational technologies~\cite{orenstrakh2024detecting, zamfirescu2023johnny}. In this study, we introduce a novel method that leverages an LLM-based agent to uncover and document tacit knowledge within organizational contexts. To validate our approach, we allow the agent to navigate virtual company hierarchies, engaging with simulated employees—also modeled as LLMs—to iteratively gather and refine details about the structure, purpose, and possible applications of a data table, particularly in the realm of machine learning tasks.

\begin{figure}
    \centering
    \includegraphics[width=\linewidth]{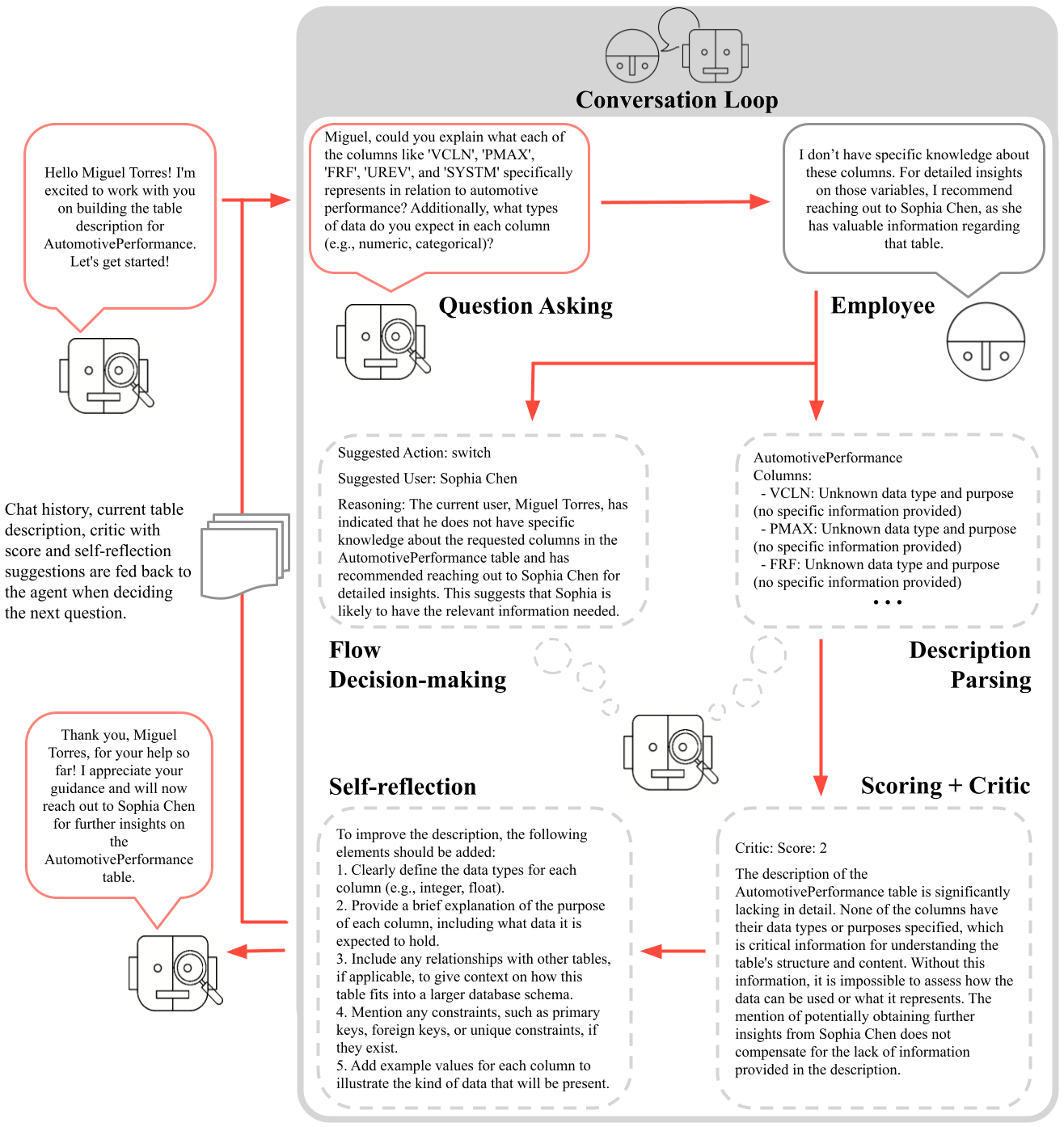}
    \caption{The agent iteratively builds knowledge and decides on its next course of action as it interacts with company employees in a conversation loop.}
    \label{fig:fluxo}
    \vspace*{-0.25cm}
\end{figure}

Our simulation-based approach reflects the complexities of real-world organizations. Employees modeled as LLMs interact with the agent, responding to inquiries, redirecting questions, or providing insights on specific aspects of the data. After each interaction, the agent assesses its progress, generating critiques and suggestions for improvement before determining the next steps as illustrated in Figure \ref{fig:fluxo}. This iterative process not only addresses knowledge gaps but also simulates human-like reasoning~\cite{webb2023emergent, miao2023selfcheck}, allowing the agent to adapt to various organizational structures and communication styles. Rather than relying on a human to locate the appropriate domain specialist within an organization—an often difficult and time-consuming task—our method facilitates the crowdsourcing of this process using LLM agents, creating a comprehensive repository of organizational knowledge.

To evaluate the effectiveness of our method, we conducted 864 simulations across diverse synthetic organizational structures, knowledge dissemination strategies, and data configurations. Our LLM agent achieved a success rate of 94.9\% in acquiring complete knowledge about the queried tables. These simulations encompassed over 300,000 interactions, including self-reasoning steps, producing a dataset exceeding 45 million words. This dataset serves as a valuable resource for advancing research into LLM-agent interactions and the relationship between organizational structures and knowledge management practices. For access to the full simulation chat log and prompts, see the Code and Data Availability statement.

%% file: Text/2related.tex
\section{Related Work}

Max Weber's seminal work on social and economic organization \cite{webertheory} offers key insights into the functioning of bureaucratic and hierarchical structures. A central focus of Weber’s work is in his concept of authority, which includes traditional authority, rooted in customs, and rational-legal authority, based on formal rules and laws. Both of these forms of authority are highly relevant to companies and economic organizations. Weber highlights bureaucracy as the primary model of rational-legal authority, emphasizing its efficiency, hierarchy, specialization, and reliance on impersonal procedures. Weber’s concept of \textit{Verstehen} (German for ``interpretive understanding") stresses the need to understand social actions through the subjective meanings individuals give to their behaviors. Additionally, his use of ``ideal types" of behavior as a methodological tool help analyze many social and economic phenomena. 
These theories form the basis for understanding the formal mechanisms that govern information flow within organizations. Many other works in organizational theory build upon Weber's ideas to explore additional aspects of organizational behavior, power dynamics, and decision-making processes. For instance, Peter M. Blau's theories of social exchange~\cite{blau2017exchange} emphasize the informal dynamics of interaction, where reciprocity and trust become critical in sharing tacit knowledge. Blau's analysis reveals how interpersonal relationships complement formal structures, resulting in collaboration and knowledge transfer.

Diefenbach and Sillince \cite{diefenbach2011formal} discuss the interplay between formal and informal hierarchies, which is crucial for understanding how knowledge propagates.  They highlight that formal hierarchies are founded on clearly delineated roles and command chains, whereas informal hierarchies emerge from social interactions and shared norms. Hedlund \cite{hedlund1993assumptions} introduced the concept of heterarchies, challenging traditional hierarchical assumptions and providing a framework to analyze knowledge flow in multinational corporations. This perspective aligns with the need to model both structured and unstructured interactions in simulations. Mihm et al. \cite{mihm2010hierarchical} explored hierarchical structures and search processes, highlighting the challenges of navigating complex organizational networks. Their study employs simulations to reveal how hierarchy impacts solution stability, quality, and speed in problem-solving, showcasing the importance of adapting organizational structures to meet search requirements effectively. 


Incorporating these social theory perspectives provides a realistic foundation for constructing synthetic company hierarchies that reflect both structured and dynamic relationships. As such, the parameter choices in our simulations are heavily influenced by these foundational works—from Weber’s concept of bureaucratic organizations to Blau’s and Diefenbach and Sillince’s emphasis on informal connections. These theoretical frameworks not only ground our models in established sociological principles, but complement the emerging computational approaches to simulating human behavior.

Recent work demonstrates LLM's ability to model complex social systems in realistic simulations. Park et al. \cite{park2023generative} proposed generative agents that simulate human behavior through dynamic memory mechanisms, enabling decision-making in a sandbox setting inspired by the game The Sims. These agents autonomously perform tasks, establish social relationships, and adapt behaviors through experiential learning.
Dai et al.~\cite{dai2024artificial} investigated LLM-driven social evolution under Hobbesian Social Contract Theory, illustrates LLMs’ capacity to simulate foundational theories of societal dynamics and demonstrating agents transitioning from conflict in a \quotes{state of nature} to cooperative social orders via emergent contractual agreements.
Qian et al.~\cite{qian2024chatdev} introduced ChatDev, a framework where role-based LLM agents collaborate through structured dialogues to complete software development phases. 
Collectively, these studies establish LLMs as tools for emulating social behaviors across individual, organizational, and societal scales.
However, to our knowledge, no previous literature proposes a method for simulating the human behaviors of interest to this work in synthetic organizational structures, namely, knowledge extraction processes.

Regarding LLM themselves, advances in prompting strategies have shown to enhance its reasoning capabilities. Prompt-chains \cite{wu2022ai} iteratively refine solutions through task-specific prompts, while least-to-most prompting \cite{zhou2022least} decomposes problems into sequential subproblems. Chain-of-thought prompting \cite{NEURIPS2022_9d560961} improves logical inference by generating intermediate reasoning steps. These divide-and-conquer approaches increase precision and reliability, particularly for iterative problem-solving contexts.
Further, techniques to introduce reasoning into LLMs have also been shown to improve results. The self-reflection paradigm~\cite{miao2023selfcheck} enables LLMs to evaluate and critique their own reasoning processes. Similarly, self-critique and self-scoring mechanisms~\cite{park2023generative} have proven effective in assessing the quality of information generated by LLMs, demonstrating the advantages of feedback and iterative refinement. Building on these strategies, our approach develops a robust agent system to extract tacit knowledge from human specialists, consolidate it into documents, and validate its efficacy. These works also show how LLM-based simulations closely mimic human interactions, making them a suitable surrogate for testing and refining such methods.

%% file: Text/3method.tex
\section{Proposed Approach}
\label{sec:approach}

The challenge of documenting tacit knowledge within an organization requires a process that accounts for both incomplete initial knowledge and the dynamic nature of organizational networks. In our proposed approach, we introduce an agent-based framework designed to gather and document knowledge about dataset tables through iterative interactions with employees. The agent starts with a limited understanding and progressively refines its knowledge by engaging with the organization. However, real-world case studies can be difficult to obtain. To evaluate the effectiveness of this methodology, we employ simulations of organizational knowledge dissemination, modeling both formal and informal networks. In this section, we describe the key aspects of our agent design and the simulations conducted to assess its performance.


\subsection{Prompt-chaining to build an Agent} \label{sec:conversation_loop}

Our approach employs a prompt-chaining technique to develop an LLM-based agent for documenting tacit knowledge. Specifically, we wish to build a description for a table known to the organization. The agent operates in a partially observable environment. While such an environment holds access to the current table description (the true knowledge \( k^* \)), each employee's expertise and network connections are uncertain. The agent does, however, know a subset of the network: the company’s hierarchical structure. This scenario can be modeled using a framework inspired by a Markov Decision Process (MDP). Briefly, it consists of:

\begin{itemize}
    \item \textbf{Knowledge States (\( \mathcal{K} \)):} Represent the agent’s current understanding of the knowledge domain.
    \item \textbf{Actions (\( \mathcal{A} \)):} Consist of the set of Actions that an agent can take.
    \item \textbf{Transition Model (\( P(k' | k, a) \)):} Probabilistic changes in knowledge state due to the actions made and changes in the environment.
    \item \textbf{Reward Function (\( R(k, a) \)):} Measures the completeness and accuracy of the knowledge state given the actions.
\end{itemize}

Formally, let \( \mathcal{K} \) represent the set of possible states that define the agent’s current knowledge level, and \( \mathcal{A} \) represent the set of actions or prompts the agent can take. The transition probability, \( P(k' | k, a) \), defines the likelihood of transitioning from state \( k \) to state \( k' \) after taking action \( a \). The reward function, \( R(k, a) \), quantifies the benefit of the action in a particular state, reflecting the knowledge acquired. The process terminates when the agent reaches a terminal state \( k_t \), where all relevant knowledge about the proposed application is acquired. This is measured by \( R(k_t, a) \geq \epsilon \), where \( \epsilon \) represents the minimum score for a suitable table description. The agent’s goal is to maximize the accumulated reward by formulating an optimal policy \( \pi^* \), which guides its decisions implicitly.

The agent begins with a basic understanding \( k_0 \) of the table or organizational knowledge, knowing only the table’s name and columns. As the agent interacts with employees, it refines its state \( k \) by incorporating new information and identifying gaps. The actions (\( \mathcal{A} \)) involve generating questions aimed at filling these gaps. The agent uses its current state to formulate questions that address specific gaps in its understanding. Additionally, the agent’s self-critical assessments, which include quality scores, explanations, and suggestions for improvement, guide it in this task. The Transition Model (\( P(k' | k, a) \)) accounts for the probabilistic nature of state transitions after an action is taken. Due to the variability in LLM outputs and employee responses, these transitions are non-deterministic. The LLM manages these uncertainties by dynamically adjusting itself and its strategies in response to the feedback it receives. The Reward Function (\( R(k, a) \)) evaluates actions, measuring alignment between the agent’s understanding and the expected complete knowledge.

While these components are not explicitly modeled in the agent framework, they describe how the LLM organizes and selects prompts. The key goal is to maximize the accumulated reward, which helps guide the agent toward more effective knowledge acquisition. The agent interacts with employees based on their position in the hierarchy, structured as a stack, and prefers to start with those at lower levels to avoid burdening their superiors. However, if an employee is mentioned during a conversation - for instance, someone who likely holds relevant information about the table - the agent updates its knowledge of the company structure and moves this person to the top of the stack. Once someone with at least partial knowledge is found, the agent can move from one recommended employee to another, severely trimming the search space. The entire process is structured as a sequence of prompts and tasks:

\begin{enumerate}[label=(\alph*)]
    \item Greet the employee to establish rapport;
    \item Formulate a question about the data table;
    \item Process the employee's response and update self internal knowledge, building a new table description;
    \item Critique the new updated description through:
    \begin{enumerate}[label=(d.\arabic*)]
        \item Scoring the new description to evaluate its quality;
        \item Criticizing it to identify gaps or inconsistencies;
        \item Suggesting areas for improvement.
    \end{enumerate}
    \item Decide whether to continue the conversation with the current employee or switch to another.
\end{enumerate}

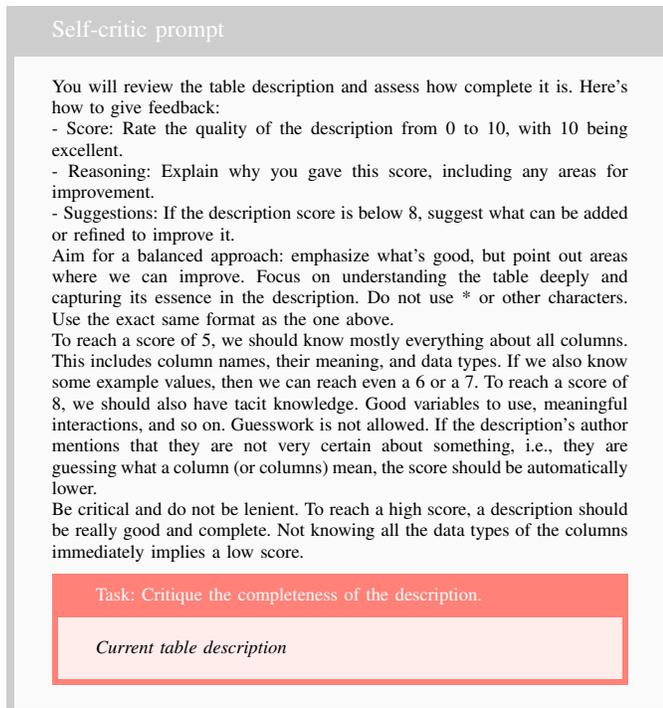
\begin{figure}[!b]
    \centering
    \small
\input{Prompts/selfcritique}
\caption{Meta-prompt used in the self-critical stage. The agent critically assesses the current table description while providing suggestions on aspects to improve.}
    \label{fig:prompt}
\end{figure}

A core aspect of this process is the self-critical feedback loop. After each interaction, the agent integrates the updated description, critique, score, suggestions, and chat history as inputs for the next question-generation step. These inputs, combined with the agent’s internal knowledge, guide the formulation of the next question, focusing on areas in which the understanding is lacking. This step involves using a meta-prompt to assess the completeness and relevance of the acquired information and evaluate its current knowledge state critically. Figure \ref{fig:prompt} illustrates the prompt used during this self-critical step. The agent then proceeds to self-reflect over the generated critic in order to evaluate the next course of action.


\subsection{Simulating company structures} \label{sec:simulate_company}

We simulate the dissemination of tacit knowledge across a synthetic company network using epidemic models~\cite{kermack1927contribution}. Specifically, we employ the Susceptible-Infectious (SI) model with waning infectivity~\cite{ehrhardt2019sir, ahmetolan2022susceptible}. In this context, knowledge about a dataset table is treated as a collection of \quotes{diseases}, where each column of the table represents a distinct \quotes{fact} (a unit of information). The dynamics of the SI model in this framework are defined as follows:

\begin{itemize} 
\item \textbf{S (Susceptible)}: Individuals who are unaware of a specific fact in the table. 
\item \textbf{I (Infectious)}: Individuals who know a fact and can share it. The likelihood of sharing decreases with time. 
\end{itemize}

The dissemination process begins with a \quotes{patient zero}, an individual who possesses complete knowledge (i.e., all facts corresponding to all columns of the table). Knowledge spreads through the organizational network, which consists of two components. The Hierarchy Network represents the formal organizational structure, explicitly known to the LLM agent; and the Relationship Network is a hidden network of informal connections (e.g., frequent interactions among colleagues) that the agent infers through its interactions. Intuitively, employees who work under the same team know each other, which leads to the Hierarchy Network being a sub-graph of the more complete Relationship Network.

Before allowing the LLM to traverse the network, we disseminate knowledge using the epidemic model across the complete Relationship Network. The key feature of this model is the waning infectivity, where an individual's ability to share knowledge diminishes over time. 
To simulate this, the time-dependent transmission rate $\beta(t)$ is modeled as:
\begin{equation}
\beta(t) = \beta_0 e^{-\gamma t}
\end{equation}

\noindent where $\beta_0$ is the initial transmission rate, and $\gamma$ determines how quickly the ability to share knowledge wanes. This SI model is used to simulate the uneven distribution of knowledge within the relationship network. Since the ability to transmit knowledge quickly wanes, over time some employees will know specific facts about the dataset table (i.e., become \quotes{infected}), while others will remain unaware. As such, we consider both:

\begin{itemize}
    \item \textbf{Knowledge Transfer}: Susceptible individuals S become informed about a fact at a rate proportional to $\beta(t) SI$, where $\beta(t)$ decays over time.
    \item \textbf{Knowledge Loss}: Infected individuals gradually lose their ability to spread a specific fact as $\beta(t)$ decreases.
\end{itemize}

The agent's task is to traverse this network after convergence using the initially incomplete knowledge about the evaluated table and employee relationships, hold conversations, and reconstruct the full table documentation by piecing together the facts that have spread across different individuals. To simulate diverse organizational structures, we define key parameters that allow us to model companies ranging from startups to large multinational corporations, from highly formal bureaucracies to agile, informal networks. 

\begin{itemize}
    \item \textbf{Max Hierarchical Depth}: Shallow hierarchies represent organizations inspired by Taylor’s principles of functional management, emphasizing task specialization, direct oversight, and higher interconnectivity among employees. These structures are often seen in manufacturing setups or flat startups. Deep hierarchies reflect Weber’s bureaucratic organization theory, characterized by formal authority, strict adherence to rules, and clear chains of command. These structures reduce interconnectivity, with knowledge flow restricted to well-defined channels. By adjusting the hierarchical depth, we can simulate a spectrum of companies, from flat startups fostering innovation to rigid bureaucracies prioritizing stability and control.
    \item \textbf{Number of Employees}: This parameter spans small teams to large multinational corporations. Smaller organizations model the dynamics of startups, where knowledge is often centralized or informally shared. Larger organizations allow us to explore the complexities of scale, including silos, multi-layered decision-making, and formalized processes. We assume balanced tree-like structures as the hierarchy. Thus, we infer the branching factor $b_d$ (that is, the number of people working under each boss) as:
    \begin{equation}
        b_d\approx \sqrt[hierarchy\text{ }depth]{ number\text{ }of\text{ }employees}
    \end{equation}
    \item \textbf{Alpha and Decay}: These parameters govern how knowledge spreads and wanes in the relationship network. Slow decay mimics organizations with stable knowledge-sharing practices. High decay represents environments where knowledge quickly becomes obsolete, such as fast-paced industries or competitive workplaces. In our simulation, decay is directly tied to how far the knowledge can spread from the initial patient zero, while alpha governs the probability of knowledge being shared.
    \item \textbf{Number of Informal Connections}: Informal connections mimic agile or networked organizations with significant ad-hoc collaboration, enabling knowledge flow outside formal channels. A low number of informal connections represents traditional, hierarchical organizations, where interactions align strictly with reporting structures. By varying the balance between formal and informal connections, we can simulate environments from rigid bureaucracies to dynamic, innovation-driven teams.
\end{itemize}

These parameters enable the simulation of various company types, including startups (flat, highly interconnected structures with informal knowledge-sharing dynamics), large multinationals (deep hierarchies, rigid formal connections, and slow diffusion of knowledge), or even agile organizations (high mean degrees, shallow hierarchies, and dynamic knowledge-sharing networks). As per Max Weber’s social theory, these simulations capture the dichotomy between traditional bureaucracies (emphasizing stability and control) and more modern, flexible organizations. 




%% file: Prompts/selfcritique.tex
{
\small
\begin{tcolorbox}[
        colback=fundo!10,        
        colframe=fundo!100,       
        sharp corners,          
        title=Self-critic prompt,
        boxrule=1mm,            
        width=\linewidth,    
        arc=4mm,                
    ]
    \scriptsize
You will review the table description and assess how complete it is. Here's how to give feedback:

- Score: Rate the quality of the description from 0 to 10, with 10 being excellent.

- Reasoning: Explain why you gave this score, including any areas for improvement.

- Suggestions: If the description score is below 8, suggest what can be added or refined to improve it.

Aim for a balanced approach: emphasize what's good, but point out areas where we can improve. Focus on understanding the table deeply and capturing its essence in the description.
Do not use * or other characters. Use the exact same format as the one above.

To reach a score of 5, we should know mostly everything about all columns. This includes column names, their meaning, and data types.
If we also know some example values, then we can reach even a 6 or a 7.
To reach a score of 8, we should also have tacit knowledge. Good variables to use, meaningful interactions, and so on.
Guesswork is not allowed. If the description's author mentions that they are not very certain about something, i.e., they are
guessing what a column (or columns) mean, the score should be automatically lower.

Be critical and do not be lenient. To reach a high score, a description should be really good and complete. Not knowing all the data types of the columns immediately implies a low score.

        \begin{tcolorbox}[
            colback=datadescription!10,        
            colframe=datadescription!70,       
            sharp corners,          
            boxrule=0.8mm,          
            arc=0mm,                
            title={Task: Critique the completeness of the description.},
            width=\linewidth,       
        ]
            \textit{Current table description}
        \end{tcolorbox}
    \end{tcolorbox}
}

%% file: Text/4exp.tex
\section{Experiments}

\begin{figure}
\begin{subfigure}[b]{\linewidth}
    \centering
    \includegraphics[height=3.4cm, width=\linewidth]{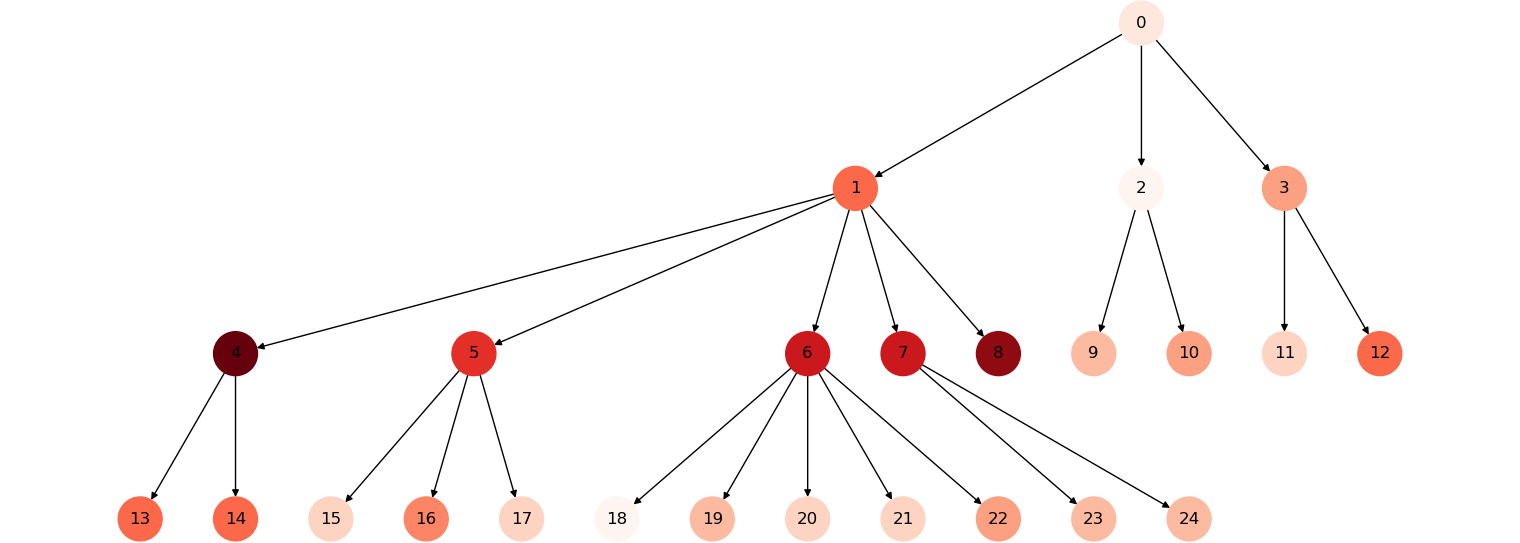}
    \caption{Hierarchy and knowledge levels.}
\end{subfigure}
\begin{subfigure}[b]{\linewidth}
    \centering
    \includegraphics[height=3.4cm, width=\linewidth]{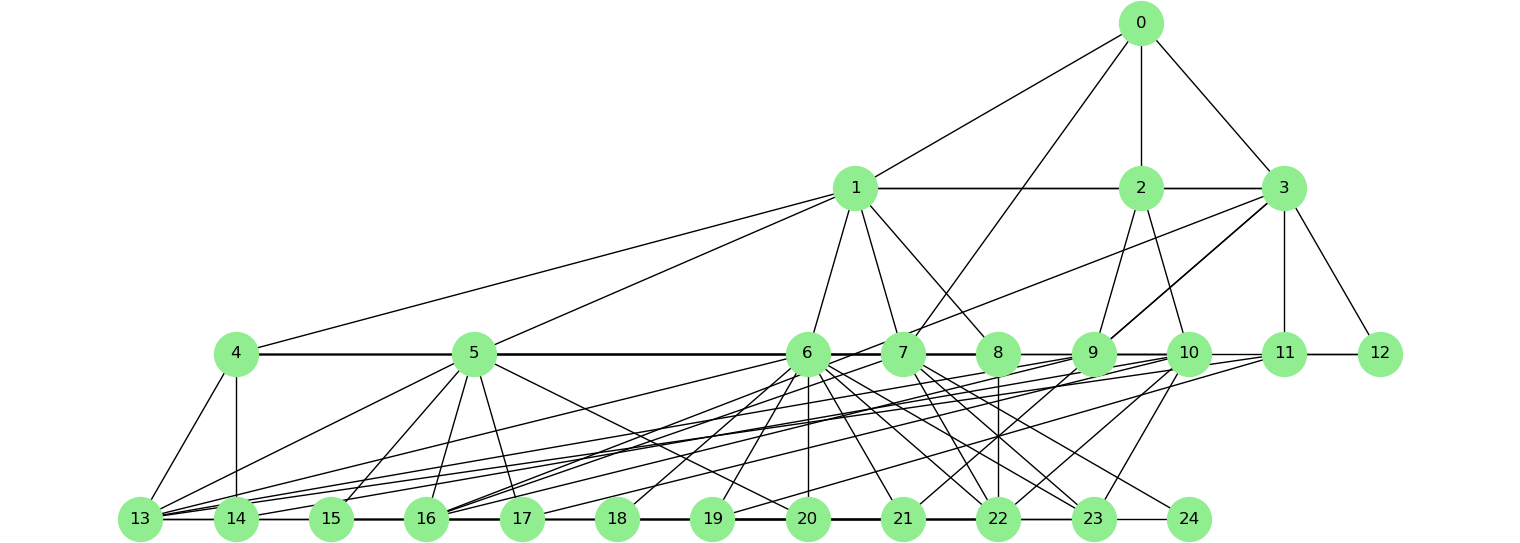}
    \caption{Relationships graph.}
\end{subfigure}
    \caption{A synthetic organization with 25 employees, 4 hierarchical steps and 2.5$\times$ informal connections. Red shades indicate the amount of knowledge (e.g.: number of columns known) that an employee holds. Decay 0.8, alpha 0.1.}
    \label{fig:company}
\end{figure}

To evaluate the effectiveness of our approach, we conducted experiments with various configurations of company structures, knowledge, and dissemination rates. Figure \ref{fig:company} illustrate an example of a synthetic organization. For each of the parameters discussed, we elected a range of possible values and iterate over all possible combinations. We assess the statistical significance of our measurements through a pairwise t-test with p-value $\leq0.05$ over 3 repetitions of each simulated environment and agent's interaction. In particular, we consider:

\begin{itemize}
\item Hierarchical Depth: 2, 5, 10, 20 (from shallow to deep)
\item Number of Employees: 20, 75, 200 (small, medium, and large organizations)
\item Alpha: 0.1, 0.5
\item Decay: 0.5, 0.8
\item Number of Informal Connections: 0, 2.5, 5 (from formal to heavily informal)
\item Number of Table Columns: 5, 20
\end{itemize}

For each repetition, a table subject was randomly selected from a predefined list, which was then used to generate the simulated complete knowledge $s^*$ through an LLM. This includes the overall table description, each column (with the respective type, meaning, and examples), as well as its possible variable relations. The possible table subjects for the generated knowledge include: aerospace, agriculture, automotive, business, construction, finances, food service, healthcare, machinery, mining, oil, packaging, energy, retail, sports, transportation, tourism, and tech. These subjects were selected following common industry sectors, as defined both in literature~\cite{arent2015key, laumas1975key, guibert1971essay} and by the International Labour Organization\footnote{\url{https://www.ilo.org/industries-and-sectors}}.

Given a parameter configuration, we then build the respective company structure, including the Hierarchy Network and the Relationship Network. In sequence, the individual facts from the original knowledge $k^*$ are disseminated in the company network using the SI epidemic model throughout the constructed company. These steps follow the methodology described in Section \ref{sec:simulate_company}. We also generate a textual background description for each employee, according to a job role as defined by the hierarchy and a randomized personality archetype, as well as a description of the information relevant to the problem they have access to: what partial knowledge of the table the individual has, their connections in the company's network, and what partial knowledge they are aware their connections have (determined probabilistically). These backgrounds are intended to make responses more human-like and also serve as context to answer the main agent's questions during the simulation. 
Finally, using the built structures and generated descriptions, we begin the conversation as described in Section \ref{sec:conversation_loop} by assigning the questioning agent a random starting employee from the bottom of the hierarchy. All experiments were conducted on OpenAI's \textit{GPT-4o mini} model.

%% file: Text/5discussion.tex
\section{Discussion}


After conducting repeated conversational experiments with varying fictional company structures, we summarized the conversational statistics by averaging the results for each company structural parameter.
We collected the self-critical score values generated during the conversations and other aspects that describe how the agent navigated the companies' internal structures.
In Table~\ref{tab_conversation_stats}, \textbf{Len(path)} represents the average number of non-unique fictitious employees contacted by the agent during inquiries about the data. In some cases, the agent contacted certain employees multiple times, effectively making these simulated individuals act as hubs within the conversational graph.
We measured the average number of hubs (\textbf{\# Hubs}) observed during the simulations.
Additionally, we calculated the minimum distance between the first person contacted by the agent and \quotes{patient zero}, who possesses all the knowledge about the table's columns. In the table, this metric is labeled \textbf{Min. dist.} and was obtained using the Dijkstra algorithm~\cite{cormen2022introduction} applied to the relationship graph of each simulated company.
We measured the percentage of times the conversational agent reached \quotes{patient zero} (\textbf{\% p0}) and reported the average and mean self-critical score produced. 

\begin{table}[!b]
    \centering
    \caption{Overview of statistics grouped and averaged for different simulated companies' parameters. NIC: number of informal connections; MHD: maximum hierarchical depth.}
    \label{tab_conversation_stats}
    \resizebox{\linewidth}{!}{
    \begin{tabular}{@{}crrrrrrr@{}}
        \toprule
        \textbf{Parameter} & \multicolumn{1}{l}{\textbf{Value}} & \multicolumn{1}{l}{\textbf{Len(path)}} & \multicolumn{1}{l}{\textbf{\# Hubs}} & \multicolumn{1}{l}{\textbf{Min. dist.}} & \multicolumn{1}{l}{\textbf{\% p0}} & \multicolumn{1}{l}{\textbf{Avg. score}} & \multicolumn{1}{l}{\textbf{Final score}} \\ \midrule
        \multirow{2}{*}{Alpha} & 0.1 & 31.6 & 1.26 & 3.94 & 0.38 & 5.49 & 6.79 \\
         & 0.5 & 16.0 & 0.59 & 3.95 & 0.22 & 6.34 & 8.07 \\ \midrule
        \multirow{3}{*}{NIC} & 0.0 & 36.34 & 1.32 & 5.89 & 0.41 & 5.35 & 6.79 \\
         & 2.5 & 19.69 & 0.82 & 3.06 & 0.26 & 6.1 & 7.66 \\
         & 5.0 & 15.38 & 0.65 & 2.89 & 0.23 & 6.3 & 7.84 \\ \midrule
        \multirow{2}{*}{Decay} & 0.5 & 27.81 & 1.06 & 3.91 & 0.34 & 5.67 & 7.2 \\
         & 0.8 & 19.8 & 0.79 & 3.99 & 0.25 & 6.16 & 7.66 \\ \midrule
        \multirow{4}{*}{MHD} & 2 & 13.81 & 0.65 & 2.72 & 0.24 & 6.37 & 7.97 \\
         & 5 & 25.0 & 1.01 & 4.01 & 0.32 & 5.8 & 7.36 \\
         & 10 & 27.85 & 1.06 & 4.39 & 0.29 & 5.74 & 7.19 \\
         & 20 & 28.56 & 0.98 & 4.66 & 0.33 & 5.75 & 7.21 \\ \midrule
        \multirow{3}{*}{\# Employees} & 20 & 8.66 & 0.6 & 2.77 & 0.48 & 6.06 & 7.57 \\
         & 75 & 26.67 & 1.08 & 4.05 & 0.28 & 5.94 & 7.59 \\
         & 200 & 36.09 & 1.1 & 5.02 & 0.14 & 5.74 & 7.13 \\ \midrule
        \multirow{2}{*}{\# Features} & 5 & 25.13 & 0.91 & 4.01 & 0.28 & 6.2 & 7.6 \\
         & 20 & 22.48 & 0.95 & 3.88 & 0.31 & 5.64 & 7.26 \\ \bottomrule
    \end{tabular}}
\end{table}




Regarding the parametization of the SI model, we observe that the average conversational path length tends to decrease as the alpha value increases, while the description scores tend to improve. Analogously, the number of non-unique employees contacted by the conversational agent decreases as the decay of knowledge transmission increases. The same decrease is observed for the number of hubs within the organization. These observations suggests that knowledge is more evenly distributed within the companies' structure, which aligns with our expectations. There does not appear to be a clear relationship between changes in the alpha value and the minimum distance to patient zero.
However, patient zero becomes less likely to be reached as the alpha value increases and the agent has more options to explore before reaching the employee holding all the information about a table.
Additionally, the number of hubs decreases with an increase in alpha.

As expected, the average path length and the distance to \quotes{patient zero} increase as the maximum hierarchy depth increases.
This indicates that the agent needs to contact more employees to retrieve the desired knowledge.
Additionally, the overall description score decreases slightly with an increase in maximum hierarchy depth, hinting at knowledge fragmentation within the company structure. On the other hand, the score values remain relatively stable despite changes in the number of employees, highlighting the robustness of the description extraction procedure regardless of company size.

We likewise observe that a higher Number of Informal Connections (NIC) corresponds to a higher self-critical score obtained at the end of the simulations. Additionally, as the NIC increases in the simulated company structure, the number of conversational hubs and the overall conversation path length decrease. Similarly, the minimum distance between the starting employee and patient zero decreases as the number of connections increases. These results are expected since a higher NIC increases the likelihood of quickly reaching employees who hold specific information about the data at hand. The number of times the employee with all the information is contacted decreases with a higher NIC, suggesting that information is more likely to be distributed among employees when they are more interconnected.
However, this observation, constrained by our simulation setup, only accounts for the target knowledge sought by the conversational agent. In real-world scenarios, informal connections are more likely to imply shared informal knowledge rather than business knowledge.







Apart from a slight decrease in score quality as the number of features increases, varying the number of features does not strongly correlate with other experimental parameters.
This is expected, as these parameters primarily influence the companies' structure rather than the knowledge retrieval task.


After analyzing the simulated conversations through which the agents reconstructed the original knowledge, we also evaluated the quality of the reproduced table descriptions that were generated at the end of each experiment run. This helps us to consider whether the proposed methodology can effectively achieve its desired goal of recovering the disseminated knowledge. For that purpose, we compare the final description $k_t$ generated by the agent with the original knowledge description $k^*$ for that table according to the following metrics:

\paragraph{Full-knowledge recall}

We first checked if the agent was able to generate a table description that included a mention of every column originally disseminated in the company's network. An agent is considered to have achieved full-knowledge recall in an experiment run if it succeeded in at least reproducing all table columns in its final report. 

\paragraph{METEOR}

We also measure the METEOR score \cite{lavie2007meteor} of the descriptions for each column in the table. The score creates word alignments between the candidate and reference texts, using not only exact matches, but also stems and WordNet synonyms and calculates the harmonic combination of precision and recall for those alignments. The METEOR scores for all columns are averaged to arrive at an overall score \textbf{cMETEOR} for the generated description $k_t$.

\paragraph{G-Eval}

We then used an LLM-as-judge approach to better capture the semantic equivalency between the original and generated descriptions. For that, we apply the G-Eval LLM evaluation framework \cite{liu2023gevalnlgevaluationusing} to score each column description according to prompted definitions of textual quality. The G-Eval framework uses LLMs with Chain-of-Thought to fill out a form and generate a final score between 1 and 5 according to the given definition. The G-Eval scores for all columns are averaged to arrive at an overall score
for the generated description $k_t$, which we refer to as \textbf{cGEvalCoh}, for a definition of ``coherence'', and \textbf{cGEvalFaith}, for a definition of ``faithfulness''.

\paragraph{Self-critical with context}

Finally, to directly compare how having access to the official table $k^*$ can change the agent's own evaluation, we altered the self-critic LLM prompts to give the evaluation score for the same final description $k_t$, but now having access to the original description as context.

From the 864 simulations executed, the LLM agent achieved full-knowledge recall in 820 of them (94.9\%). As such, in the vast majority of cases, the agent at a minimum retained some mention of every column present in the original table $k^*$. In the remaining 44 cases, the final description $k_t$ retained $\sim77\%$ of the expected columns on average.



The agent-generated descriptions measured on average $0.17$ on the column-based METEOR score, $2.65$ for column-based G-Eval Coherence and $4.37$ for G-Eval Faithfulness. In addition, adding the original description to the self-critical evaluation reduced the average score from $7.43$ to $6.75$. While looking directly at the value of those metrics is insufficient to evaluate the quality of the created texts definitively, we can analyze their values in relation to each other and to additional attributes of the simulation runs. 

Table~\ref{tab_evaluation} shows the rank correlation between the final score given by the self-critic agent during the simulation and the evaluation metrics based on the original knowledge. We identify a strong correlation between the self-critical score with and without context, indicating that the agent's own evaluation during the conversation partly maintains its relative ranking once it gains access to the original knowledge, even if the absolute score does get reduced. We also observe that the cMETEOR and cGEvalCoh metrics are strongly correlated, indicating that both the word alignment-based metric and the LLM-as-judge method similarly measure the correspondence between the target and reference texts in this context. cGEvalFaith also showed a moderate correlation with the other evaluation metrics, such as $0.66$ to the score with context.

\begin{figure}
    \centering
    \subfloat[cGEvalFaith]{\includegraphics[width=0.48\linewidth]{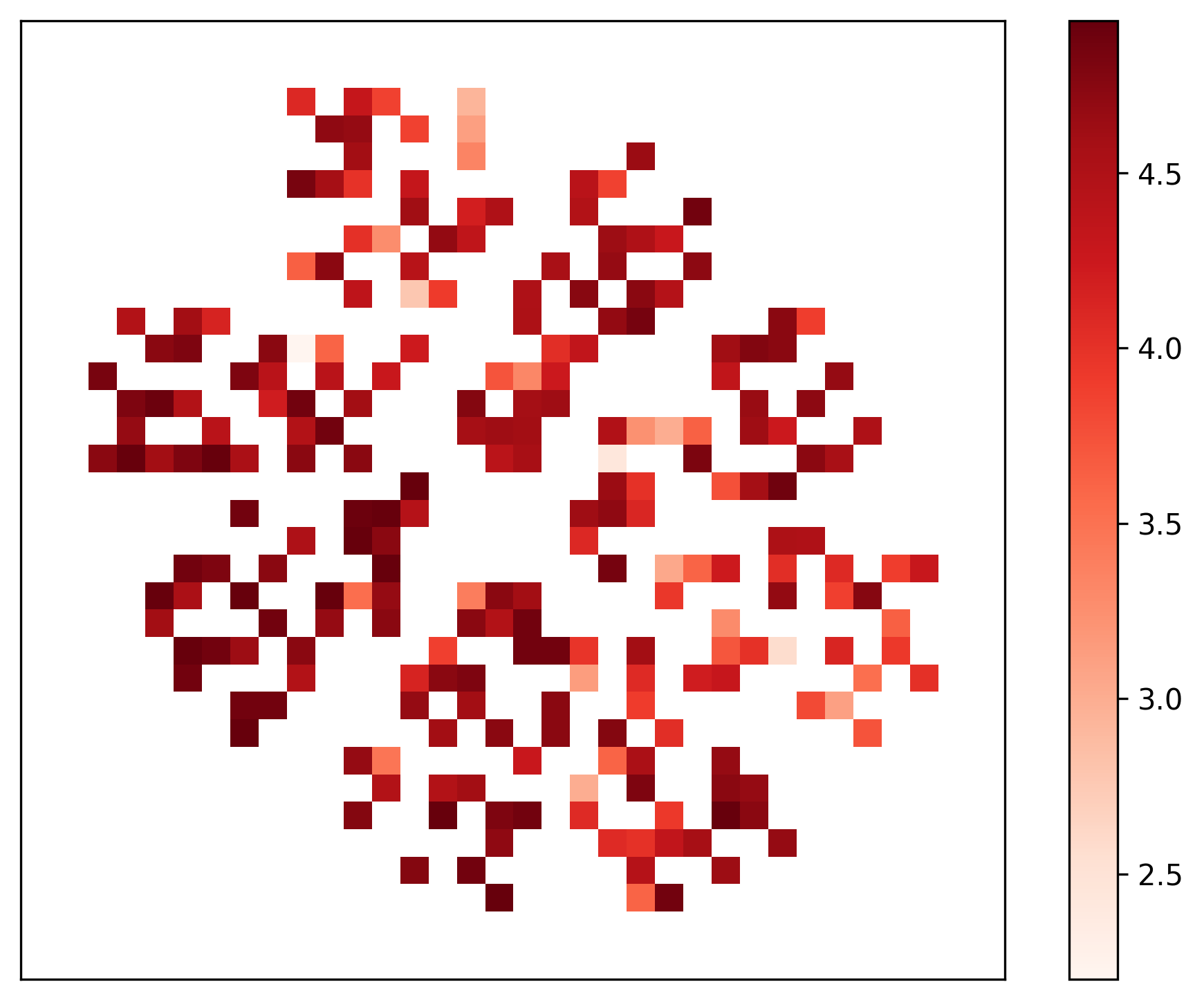}}
    \hspace{0.1cm}
    \subfloat[\% Patient $p_0$ reached]{\includegraphics[width=0.48\linewidth]{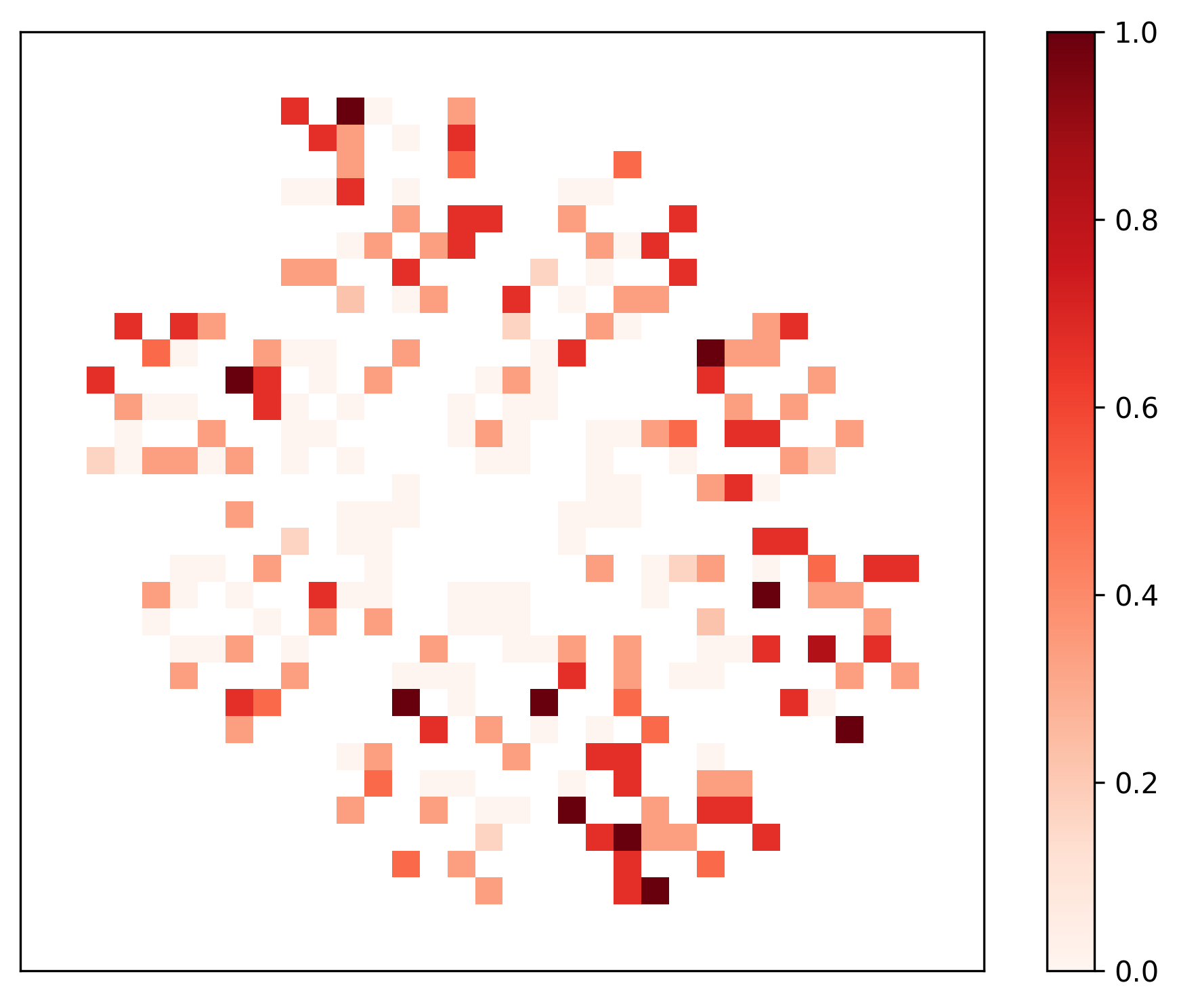}}
    
    \caption{UMAP comparing the quality of the obtained table descriptions against the dependency on contacting patient zero.}
    \label{fig_dep_p0}
\end{figure}

Lastly, we assess the effectiveness of our proposed autonomous agent-based knowledge retrieval process by evaluating how the quality of the obtained descriptions is affected by interacting with patient zero or not. 
In our simulated setup, once the agent reaches patient zero, the task becomes trivial. Ideally, the agent should be able to retrieve high-quality descriptions from employees who do not hold full knowledge. We use the company simulation parameters discussed in Table~\ref{tab_conversation_stats} as inputs for dimensionality reduction with the UMAP algorithm~\cite{mcinnes2020umapuniformmanifoldapproximation} and aggregated points over a grid with colors representing either cGEvalFaith or \% p0 (averaged over the three repetitions performed for each configuration).
The resulting projections are shown side by side in Figure~\ref{fig_dep_p0}.
We identify areas with high-quality descriptions where patient zero was never contacted (e.g., the center region of the projected points).
Conversely, we also observe areas where patient zero was contacted and the obtained cGEvalFaith was high.
These findings, along with the low correlation ($-0.06$) between the compared metrics, reinforce that the effectiveness of the proposed agent does not depend on reaching the employee $p_0$ who possesses complete knowledge.

\begin{table}
    \setlength{\tabcolsep}{3.5pt}
    \centering
    \caption{Spearman correlation between metric pairs and metrics averaged over the simulations (bottom row). \textbf{SCS:} self-critical score; \textbf{SCS+C:} self-critical score with context.}
    \label{tab_evaluation}
    \resizebox{\linewidth}{!}{
    \begin{tabular}{@{}p{0.2\linewidth}ccccc@{}}
        \toprule
        {} & \textbf{cMETEOR} & \textbf{cGEvalCoh} & \textbf{cGEvalFaith} & \textbf{SCS} & \textbf{SCS+C} \\ \midrule
        \textbf{cMETEOR}                &  -- &   0.772820 &    0.436230 &           0.263699 &                0.279494 \\  
        \textbf{cGEvalCoh}              &  0.772820 &   -- &    0.565856 &           0.347426 &                0.447342 \\  
        \textbf{cGEvalFaith}            &  0.436230 &   0.565856 &    -- &           0.497946 &                0.665396 \\  
       \textbf{SCS}      & 0.263699  & 0.347426  & 0.497946  & --       & 0.729141 \\ 
        \textbf{SCS+C} & 0.279494  & 0.447342  & 0.665396  & 0.729141 & --       \\ \midrule 
        \textbf{Mean Overall Score}          &  \multirow{2}{*}{0.1741} &   \multirow{2}{*}{2.6531} &  \multirow{2}{*}{4.3675}  &    \multirow{2}{*}{7.4317}      &   \multirow{2}{*}{6.7465}  \\
        \bottomrule
        \end{tabular}
    }
\end{table}

%% file: Text/6conclusion.tex
\section{Conclusion}

In this paper, we propose a novel approach for tacit knowledge retrieval using LLM-based agents within organizational environments. We map the problem as a task to reconstruct database table descriptions that have been disseminated through a company's employee network. Our approach allows the agent to navigate through the company graph, engaging in natural language conversations with employees to gradually accumulate partial information of the desired table and gain context to help direct its next course of action.

To validate our proposed method, we explore a broad simulated setup that encompasses a diverse range of companies structures with different generated table descriptions. Our empirical findings show that the proposed approach is robust and effective in retrieving tacit knowledge spread within the hierarchy of the simulated companies. By evaluating the reference-free self-critical scores used by our agent during its exploration process, we observe that these scores exhibit similarities to the reference-based evaluation metrics considered in our setup. We also identify that the agent is often able to retrieve the full table description without ever directly contacting the employee that was the source of the disseminated knowledge, achieving a recall rate of $94.9\%$. These results showcase the robustness of our approach and its ability to reconstruct tacit knowledge through automated conversational interactions.
For future work, we are implementing our approach at Kunumi, letting the agent interact with real employees in the company network. While further research is needed, preliminary results indicate improved documentation quality, enhancing workflows and efficiency as knowledge concentrated among managers becomes more accessible.